%% file: bare_conf.tex
\documentclass[conference]{IEEEtran}

\usepackage{booktabs} 
\usepackage{subfigure}
\usepackage{multirow}
\usepackage{bm}
\usepackage{amsmath, amscd, amsthm, amssymb, mathrsfs,amsfonts}
\usepackage{graphicx}
\usepackage{color}
\usepackage{mathtools}

\input{defs}

\DeclareMathOperator*{\argmax}{argmax} 
\newcommand{\xhdr}[1]{\subsubsection*{\bf #1}}

\usepackage{enumitem}
\setlist[description]{labelindent=0mm,leftmargin=0\parindent,listparindent=\parindent,parsep=\parskip}
\setlist[itemize]{leftmargin=\parindent,listparindent=\parindent,parsep=0.5\parskip}
\setlist[enumerate]{leftmargin=\parindent,listparindent=\parindent,parsep=0.5\parskip}

\newcommand{\eq}[1]{(eq.~\ref{#1})}
\newcommand{\etal}{\emph{et al.}}
\usepackage{cite}

\ifCLASSINFOpdf
\else
\fi
\usepackage[hidelinks,bookmarks=false]{hyperref}
\hypersetup{
	bookmarks=false,
    colorlinks=true,
    linkcolor=black,
    citecolor=black,
    filecolor=black,
    urlcolor=black,
}

\begin{document}
\title{Visually-Aware Fashion Recommendation and Design with Generative Image Models}


\author{\IEEEauthorblockN{Wang-Cheng Kang}
\IEEEauthorblockA{UC San Diego\\
wckang@eng.ucsd.edu}
\and
\IEEEauthorblockN{Chen Fang}
\IEEEauthorblockA{Adobe Research\\
cfang@adobe.com}
\and
\IEEEauthorblockN{Zhaowen Wang}
\IEEEauthorblockA{Adobe Research\\
zhawang@adobe.com}
\and
\IEEEauthorblockN{Julian McAuley}
\IEEEauthorblockA{UC San Diego\\
jmcauley@eng.ucsd.edu}
}
%
%




\maketitle

\begin{abstract}
Building effective recommender systems for domains like fashion is challenging due to the high level of subjectivity and the semantic complexity of the features involved (i.e., fashion styles). Recent work has shown that approaches to `visual' recommendation (e.g.~clothing, art, etc.) can be made more accurate by incorporating visual signals directly into the recommendation objective, using `off-the-shelf' feature representations derived from deep networks. Here, we seek to extend this contribution by showing that recommendation performance can be significantly improved by learning `fashion aware' image representations directly, i.e., by training the image representation (from the pixel level) and the recommender system jointly; this contribution is related to recent work using Siamese CNNs, though we are able to show improvements over state-of-the-art recommendation techniques such as BPR and variants that make use of pre-trained visual features. Furthermore, we show that our model can be used \emph{generatively}, i.e., given a user and a product category, we can generate new images (i.e., clothing items) that are most consistent with their personal taste. This represents a first step towards building systems that go beyond recommending existing items from a product corpus, but which can be used to suggest styles and aid the design of new products.
\end{abstract}

\section{Introduction}

The goal of a recommender system is to provide personalized suggestions to users, based on large volumes of historical feedback, by uncovering hidden dimensions that describe the preferences of users and the properties of the items they consume. Traditionally, this means training predictive algorithms that can identify (or rank) items that are likely to be clicked on, purchased (or co-purchased), or given a high rating. In domains like fashion, this can be particularly challenging for a number of reasons: the vocabulary of items is long-tailed and new items are continually being introduced (cold-start); users' preferences and product styles change over time; and more critically, the semantics that determine what is `fashionable' are incredibly complex.

\begin{figure}[h]
\begin{center}
\includegraphics[width=.99\linewidth]{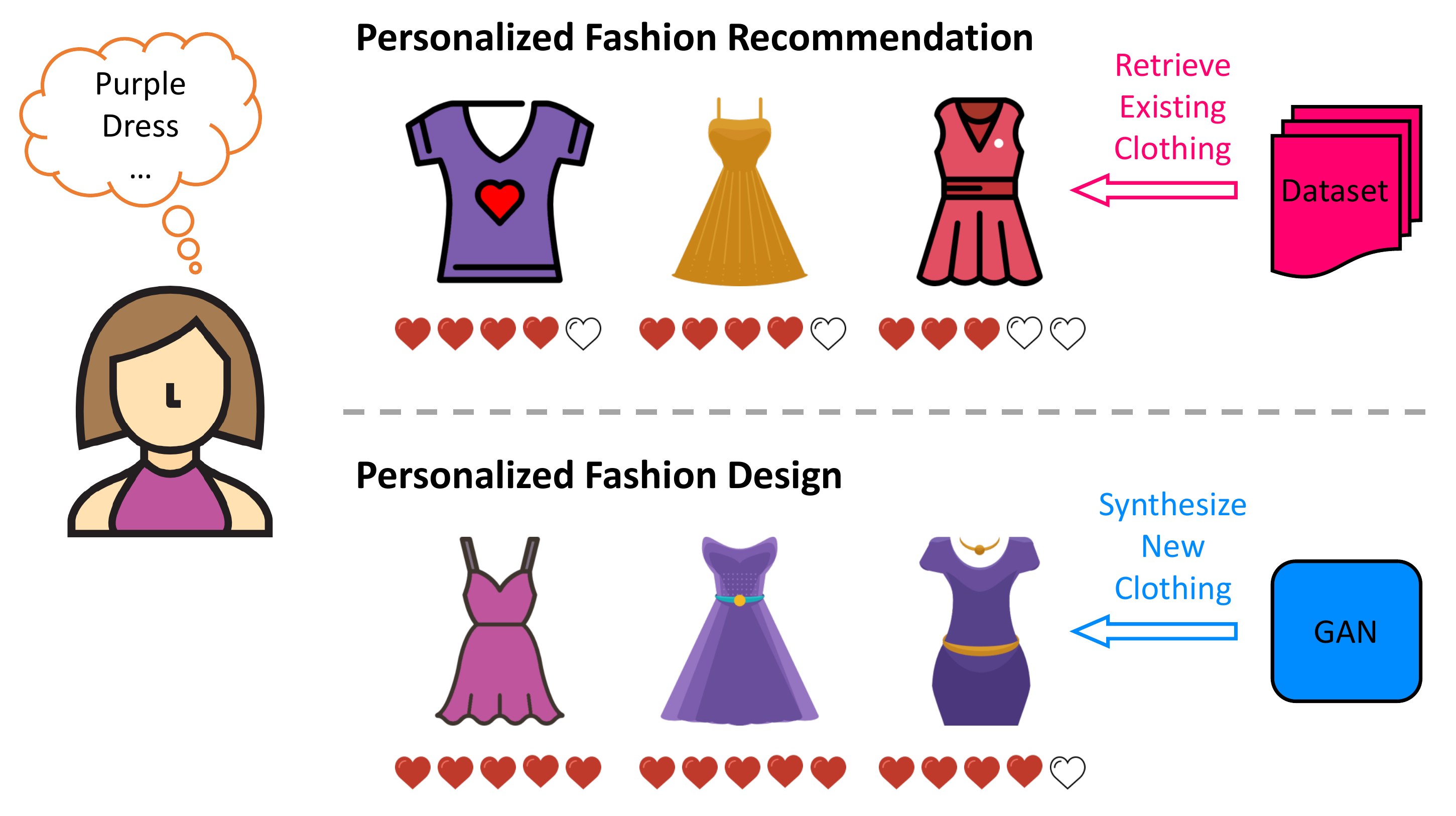}
\caption{Our model can be used for both personalized recommendation and design. Personalized recommendation is achieved by using a `visually aware' recommender based on Siamese CNNs; generation is achieved by using a 
Generative Adversarial Net to synthesize new clothing items in the user's personal style. \label{fig:intro} (Icons made by \href{https://smashicons.com/}{Madebyoliver}, \href{https://roundicons.com/}{{Roundicons}} and \href{http://www.freepik.com/}{Freepik} from www.flaticon.com)}
\end{center}
\end{figure}

Recently, there has been an effort to address these challenges by building recommender systems that are `visually aware,' i.e., which incorporate visual signals directly into the recommendation objective.
In the simplest case, visual features can be taken `off the shelf,' and incorporated into a content-aware recommender system \cite{VBPR}. Although the features captured by such methods (e.g.~CNN features extracted from Caffe \cite{CAFFE}) describe high-level characteristics (e.g.~color, texture, shape), and not necessarily the subtle characteristics of `fashion,' these methods are nevertheless effective, especially in `cold-start' settings where high-level image characteristics can be informative. Such work has been extended to incorporate temporal and social signals, and to model visual factors in other settings, like artistic recommendation \cite{upsdowns,HeFanWanMcA}.

A parallel line of work has sought to develop new image representations using convolutional neural networks. This includes image classification, 
as well as models specifically designed for fashion, e.g.~to identify compatible items by training 
on large corpora of
co-purchase dyads \cite{VeiKovBelMcABalBel15}.

\xhdr{In this paper} we seek to combine these two lines of work, by training image representations specifically for the purpose of fashion recommendation. In other words, we seek to use image content (at the pixel level) to build recommender systems. This follows the recent trend of incorporating representation learning techniques into recommender systems \cite{wang2015collaborative}, and methodologically is most similar to the work of \cite{lei2016comparative} where comparative judgments between images are modeled using a certain type of Siamese network. We adapt the popular formulation from Bayesian Personalized Ranking (BPR) \cite{rendle2009bpr} to include image content via a Siamese net and show significant improvements over BPR itself, as well as extensions of BPR that make use of pre-trained representations \cite{VBPR}.

\subsection{Toward Recommender Systems for Design} Beyond demonstrating improved recommendation and retrieval performance, our second contribution is to show that our model can be combined with Generative Adversarial Networks (GANs) \cite{goodfellow2014generative} to \emph{generate} images that are consistent with those in the training corpus, and can be conditioned on a particular user to maximize their personal objective value. In other words, given a particular user and category, we can generate a new (image of an) item that is most consistent with a user's preference model.
This idea consists of using a GAN to take
us from the discrete space of fashion images to a \emph{continuous} space, allowing us to explore the potential range of fashion items, suggest modifications to existing items, or generate items that are tailored to a specific user.

The items generated look plausible yet substantially different from those in the training corpus; this represents a first step toward using recommender systems not just to suggest existing items, but also to aid in the design of new ones.

\emph{Methodologically,} our system builds upon BPR \cite{rendle2009bpr} and Siamese networks \cite{hadsell2006dimensionality} to build a visually-aware personalized recommender system that for each user optimizes pairwise preferences between purchased versus non-purchased items, based on latent properties of the items and their product images. This allows us to generate personalized preference models and generate rankings of likely purchases. Next, we combine the learned system with a Generative Adversarial Net, so that given a user and product category, we can synthesize new items (or item images) that are maximally consistent with each user's preference model. Similar to the method of activation maximization \cite{DBLP:conf/nips/NguyenDYBC16}, our synthesis component iteratively finds a low dimensional latent code, which generates an item image that maximizes a personal objective value. 

\emph{Quantitatively,} we perform experiments on a popular fashion recommendation corpus from \emph{Amazon} \cite{julian}, which includes hundreds of thousands of users, items, and reviews. We show state-of-the-art results in terms of recommendation performance (in terms of the AUC) compared to BPR and variants that make use of product images via pre-trained features \cite{rendle2009bpr,VBPR}. We also show that our GAN architecture is able to synthesize images with substantially higher objective values for each user than existing items in the corpus. 

\emph{Qualitatively,} the clothing images generated by our system are plausible and diverse, meaning that the system can feasibly be used in an exploratory way to understand the potential space of fashion items, and to identify items that match individual users' preferences. This is in line with recent work that aims to make recommender systems more interpretable and explainable, and to generate `richer' recommendations---rather than simply recommending products from an existing corpus, we can help users and designers explore new styles and design new items.

%

The overall idea of our system is shown in Figure~\ref{fig:intro}.
We envision a system capable of performing both recommendation (retrieval) and synthesis: Our system architecture is based on the Siamese CNN framework, where the final layer of the CNN is treated as an item representation that can be used in a preference prediction framework; for image synthesis we use the framework of Generative Adversarial Nets, where a generator and discriminator are simultaneously trained so as to generate images that follow the same distribution as (and hence can't be distinguished from) those in the training corpus. We demonstrate this framework in a variety of design scenarios.





\section{Related Work}

We extend work on visually-aware recommendation with recent advances on representation learning and generative models of images. We highlight a few of the main ideas from each area below.

\xhdr{Recommender Systems}
At their core, recommender systems seek to model 
`compatibility' between users and items, based on historical observations of clicks, purchases, or interactions.
\emph{Matrix Factorization} (MF) methods relate users and items by uncovering latent dimensions such that users have similar representations to items they rate highly, and are the basis of many state-of-the-art approaches. 
We are concerned with personalized ranking from \emph{implicit} feedback (e.g.~distinguishing purchases from non-purchases, rather than estimating ratings), 
which can be challenging 
due to the ambiguity of interpreting `non-observed' (e.g.~non-purchased) data.
Recently, \emph{point-wise} and \emph{pairwise} methods have successfully adapted MF to address such challenges.

\emph{Point-wise} methods assume non-observed feedback to be inherently negative, and cast the task in terms of regression,
either by associating 
`confidence levels' to 
feedback \cite{WRMF}, or by sampling non-observed feedback as negative instances \cite{OCCF}.

\emph{Pairwise} methods are based on a weaker but possibly more realistic assumption that positive feedback must only be `more preferable' than non-observed feedback.
Such methods directly optimize the ranking (in terms of the AUC) of the feedback and are to our knowledge state-of-the-art for implicit feedback datasets. In particular, Bayesian Personalized Ranking (BPR), has
experimentally been shown to 
outperform a variety of competitive baselines \cite{rendle2009bpr}.
Furthermore, BPR-MF
(i.e., BPR with MF as the underlying predictor)
has been successfully extended to incorporate (pre-trained) visual signals \cite{VBPR}, and is thus the framework on which we build.


Besides images, others have developed content- and context-aware models that make use of a variety of information sources, 
including 
temporal dynamics \cite{DBLP:conf/icdm/YangLTXLZ15}, 
and geographic information \cite{DBLP:conf/icdm/LianGZYXZR15,DBLP:conf/icdm/YaoFLLX16}. Like our work these fall under the umbrella of `content aware' recommender systems, though are orthogonal due to the application domain and choice of signals considered.

\xhdr{Deep Learning-Based Recommender Systems}
A variety of approaches have sought to incorporate `deep' learning into recommender systems \cite{wang2015collaborative
}, including systems that make use of content-based signals, such as for recommendation of \emph{YouTube} videos \cite{covington2016deep}. Our work is also a form of deep recommender system, though is orthogonal to existing approaches in terms of data modality, except for a few notable exceptions described below.









\xhdr{Visually-aware Recommender Systems}
Recent works have introduced visually-aware recommender systems where users' rating dimensions are modeled in terms of visual signals in the system (product images). Such visual dimensions were demonstrated to be successful at link prediction tasks, such as recommending alternative
(e.g.~two similar t-shirts) 
and complementary 
(e.g.~a t-shirt and a matching pair of pants) 
items \cite{julian}, and POI recommendation\cite{DBLP:conf/www/WangWTSRL17}.

The most closely related work is a recent method 
that extended BPR-MF
to incorporate
visual dimensions to facilitate item recommendation tasks \cite{VBPR}. We build upon this framework, showing that substantially improved performance can be obtained by 
using 
an `end-to-end' learning approach (rather than pre-trained features).

\xhdr{Fashion and Clothing Style}
Beyond the methods mentioned above, modeling fashion or style characteristics has emerged as a popular computer vision task in settings other than recommendation~\cite{DBLP:journals/tmm/LiCZL17,DBLP:conf/mm/HanWJD17,DBLP:conf/iccv/Al-HalahSG17},
e.g.~with a goal to
categorize or extract features from images, without necessarily building any model of a `user.'
This includes categorizing images as belonging to a certain style \cite{bossard2013apparel}, as well as assessing items (or individuals \cite{murillo2012urban}) for compatibility \cite{runway2realwayWACV15,julian}.


\xhdr{Siamese Networks and Comparative Image Models}
Convolutional Neural Networks (CNNs) have experienced a resurgence 
due to their success as general-purpose image models, especially for classification \cite{krizhevsky2012imagenet}. 
In particular, \emph{Siamese nets} \cite{hadsell2006dimensionality} have become popular for metric learning and retrieval, as they allow CNNs to be trained `comparatively;' here two `copies' of a CNN are joined by a function that compares their outputs. This type of architecture has been applied to discriminative tasks (like face verification) \cite{hu2014discriminative}, as well as comparative tasks, such as modeling preference judgments between images \cite{lei2016comparative}. Such models have also been used to learn notions of `style' \cite{bell15siggraph}, including for fashion, by modeling the notion of item compatibility; however such systems are focused on comparison and retrieval, rather than personalized recommendation.

The closest works to ours in this line are those of Veit \etal~\cite{SiameseICCV} and Lei \etal~\cite{lei2016comparative}. The former considers item-to-item recommendation tasks on the same \emph{Amazon} dataset used here, though is concerned with learning a \emph{global} notion of compatibility and has no personalization component. The latter uses a similar triplet-embedding framework to ours, albeit for a different objective and a different domain. However neither work considers using the model \emph{generatively} for image synthesis and design as we do here.

\xhdr{Image Generation and Generative Adversarial Networks}
Generative Adversarial Networks (GANs) are an unsupervised learning framework in which two components `compete' to generate realistic looking outputs (in particular, images) \cite{goodfellow2014generative}. One component 
(a generator)
is trained to generate images, while another (a discriminator) is trained to distinguish real versus generated images. Thus the generated images are trained to look `realistic' in the sense that they are indistinguishable from those in the dataset.

Such systems can also be conditioned on additional inputs, in order to sample outputs with certain characteristics
\cite{condgan}. We follow a similar approach based on activation maximization \cite{DBLP:conf/nips/NguyenDYBC16} to generate images that best match a selected user's personal style. GANs have been used for broad applications, though to our knowledge we are the first to apply this architecture to fashion image generation.

\subsection{Key Differences}
In summary, our method 
is distinct from most recommendation approaches through the use of visual signals, and specifically
extends work on visually-aware recommendation by using richer image representations;
this leads to
state-of-the-art performance in terms of personalized ranking. 
Most novel is the ability to use our 
system \emph{generatively}, suggesting a form of recommender system that can help with the exploration and design of new items.

\section{Methodology}
Our system has two important parts: a component for visual recommendation (essentially a combination of a Siamese CNN with Bayesian Personalized Ranking), and a component for image generation (based on Generative Adversarial Networks).
Our notation is defined in Table \ref{tab:notation}.
We describe each component separately.

\begin{table}[t]
\caption{Notation. \label{tab:notation}}
\begin{tabular}{lp{4.9cm}}

\toprule
Notation&Explanation\\
\midrule
$\mathcal{U},\mathcal{I}$ & user and item set\\
$\mathcal{I}_u^+$ & positive item set for user $u$\\
$x_{u,i}\in\mathbb{R}$ & predicted score user $u$ gives to item $i$ \\
$K\in \mathbb{N}$ & latent factor dimensionality\\
$\btheta_u\in\mathbb{R}^K$& latent factor for user $u$\\
$\Phi$
& convolutional network for feature extraction\\
$D$
& discriminator in GAN\\
$G$
& generator in GAN\\
$\z\in [-1,1]^{100}$ & latent input code for the generator\\
$\Delta,\rDelta$&image upscaling/downscaling operator \\
$\X_i\in\mathbb{R}^{224\times 224\times 3}$ & product image for item $i$ \\
$X_c$ & product image set for category $c$ \\
\bottomrule
\end{tabular}
\end{table}

\subsection{Visually-Aware Recommender Systems}
We start by developing an end-to-end 
visually-aware
deep Bayesian personalized ranking method (called \emph{DVBPR}) to simultaneously extract task-guided visual features and to learn user latent factors. We first define the visually-aware recommendation problem with implicit feedback and introduce our preference predictor function. 
Afterward we outline our procedure for model training.

\xhdr{Problem Definition}
We consider recommendation problems with implicit feedback (e.g.~purchase/click histories, as opposed to ratings). 
In the implicit feedback setting, there is no negative feedback; instead our goal is to rank items such that `observed' items are ranked higher than non-observed ones
\cite{rendle2009bpr}. This approach was shown to be successful in previous work on visual recommendation \cite{VBPR}, to which we compare our results.

We use $\mathcal{U}$ and $\mathcal{I}$ to denote the set of users and items (respectively) in 
our platform.
For our implicit feedback dataset, $\mathcal{I}_u^+$ denotes a set that includes all items about which user $u$ has expressed positive feedback (e.g.~purchased, clicked). 
Each item $i\in\mathcal{I}$ is associated with an image, denoted
$\X_i$. 
Our goal is to generate for each user $u$ a personalized ranking over items with which the user $u$ has
not yet interacted.

\xhdr{Preference Predictor}
Matrix Factorization (MF) methods have shown state-of-the-art performance both for rating prediction and modeling implicit feedback \cite{korenSurvey,rendle2009bpr}. A basic predictor (that we use as a building block) to predict the preference of a user $u$ about an item $i$ is
\begin{equation}
\label{eq:standardMF}
x_{u,i}=\alpha+\beta_u+\beta_i+\bgamma_{u}^{T}\bgamma_i,
\end{equation}
where $\alpha$ is an offset, $\beta_u$ and $\beta_i$ are user/item biases, and $\bgamma_{u}$ and $\bgamma_{i}$ are latent factors that describe user $u$ and item $i$ respectively. These latent factors can be thought of as the `preferences' of the user and the `properties' of an item, such that a user is likely to interact with an item if their preferences are compatible with its properties.

Previous models for visually-aware recommendation made use of \emph{pre-trained} visual features, like  \cite{VBPR}, adding additional item factors 
by embedding the visual features into a low-dimensional space:
\begin{equation}
\label{eq:VBPR}
x_{u,i}=\alpha+\beta_u+\beta_i+\overbrace{\bgamma_{u}^{T}\bgamma_i}^{\mathclap{\text{\emph{latent} user-item preference}}}+\underbrace{\btheta_{u}^{T}(\E \f_i).
}_{\mathclap{\text{\emph{visual} user-item preference}}}
\end{equation}
Above $\btheta_{u}$ is a (latent) visual preference vector for user $u$, $\f_i\in\mathbb{R}^{4096}$ is a pre-extracted CNN feature representation of item $i$, and $\E\in\mathbb{R}^{K\times 4096}$ is an embedding matrix that projects $\f_i$ into a $K$-dimensional latent space, whose dimensions correspond to facets of `fashion style' that explain variance in users' opinions. 

The main issue we 
seek to address 
with the method above
is its reliance on \emph{pre-trained} features that were optimized for a different task (namely, classification on ImageNet).
Thus we wish to evaluate the extent to which task-specific features can be used to improve performance within an end-to-end framework. 
This builds upon recent work on comparative image models \cite{lei2016comparative,SiameseICCV}, though differs in terms of the formulation and training objective used.



To achieve this,
we 
replace the pre-trained visual features and embedding matrix with
a CNN network $\Phi(\cdot)$ to extract visual features directly from the images themselves:
\begin{equation}
x_{u,i}=\alpha+\beta_u+\btheta_{u}^{T}\Phi(\X_i).
\end{equation}

Unlike standard MF (eq.~\ref{eq:standardMF}) and VBPR (eq. \ref{eq:VBPR}), we exclude item bias terms $\beta_i$ and non-visual latent factors $\gamma_i$. 
Empirically, we found that doing so improved performance, and that the remaining terms in the model are sufficient to capture these factors implicitly.


\xhdr{Feature Extraction}
Our Convolutional Neural Network $\Phi(\cdot)$ is based on the \emph{CNN-F} architecture from \cite{chatfield2014return}. This is a CNN architecture designed specifically for efficient training. Using more powerful architectures~(e.g.~ResNet \cite{he2016deep}),
may achieve better performance; however, we found that \mbox{CNN-F} is sufficient to show the effectiveness of our method, while allowing for training on standard desktop hardware. Specifically, CNN-F consists of
8 learnable layers, 5 of which are convolutional,
while the last 3 are fully-connected. The input image
size is $224\times 224$. 


Details of our architecture are shown below (see \cite{chatfield2014return}, Table 1 for details; st=stride; pad=spatial padding):
\begin{center}
\noindent
{%
\scriptsize
\setlength{\tabcolsep}{2.2pt}
\begin{tabular}{|cccccccc|}
\hline
conv1 & conv2 & conv3 & conv4 & conv5 & full6 & full7 & full8\\
64x11x11 & 256x5x5 & 256x3x3 & 256x3x3 & 256x3x3 & 4096 & 4096 & \textbf{$K$}\\
st.~4, pad 0 & st.~1, pad 2 & st.~1, pad 1 & st.~1, pad 1 & st.~1, pad 1 & drop- & drop-  & -\\
x2 pool & x2 pool & - & - & x2 pool & out & out & -\\
\hline

\end{tabular}
\setlength{\tabcolsep}{6pt}
}
\end{center}
In our experiments, the probability of dropout is set to $0.5$, and the weight decay term is set to $10^{-3}$.

Note one difference between the above architecture and that of CNN-F: our last layer has $K$ dimensions, whereas that of CNN-F has
1000. Here, instead of seeking a final layer that can be adapted to general-purpose prediction tasks, 
we hope to learn a representation whose
dimensions explain the variance in users' fashion preferences. 


\subsection{Learning the Model}

Bayesian Personalized Ranking (BPR) is a state-of-the-art 
ranking optimization framework for implicit feedback. In BPR, the main idea is to optimize rankings 
by considering triplets $(u,i,j)\in \mathcal{D}$, where 
\[\mathcal{D}=\{(u,i,j)|u\in\mathcal{U}\wedge i\in\mathcal{I}_u^+\wedge j\in\mathcal{I}\setminus \mathcal{I}_u^+\}.\]
Here $i\in\mathcal{I}_u^+$ is an item about which the user $u$ has expressed interest, whereas $j\in\mathcal{I}\setminus \mathcal{I}_u^+$ is one about which they have not. Thus intuitively, for a user $u$, the predictor should assign
a larger preference score
to item $i$ than item $j$. Hence BPR defines
\begin{equation} 
x_{u,i,j}=x_{u,i}-x_{u,j}
\label{eq:diff}
\end{equation}
as the difference between preference scores, and seeks to optimize an 
objective function given by
\begin{equation}
\max\sum_{(u,i,j)\in \mathcal{D}} \ln\sigma (x_{u,i,j})-\lambda_\Theta\|\Theta\|^2,
\end{equation}
where $\sigma(\cdot)$ is the sigmoid function, $\Theta$ includes all model parameters, and $\lambda_\Theta$ is a regularization hyperparameter.

By considering a large number of samples of non-observed items $j\in\mathcal{I}\setminus \mathcal{I}_u^+$, this method can be shown to approximately optimize the AUC in terms of ranking observed feedback for each user \cite{rendle2009bpr}.

Note that in \eq{eq:diff}, the global bias term $\alpha$ and user bias term $\beta_u$ 
can be
discarded, as they cancel between $x_{u,i}$ and $x_{u,j}$. Hence the final form of our preference predictor 
can be simplified as
 \begin{equation}
\label{eq:DVBPR}
x_{u,i}=\btheta_{u}^{T}\Phi(\X_i).
\end{equation}

Since all parts of the objective 
are
differentiable, we 
perform optimization by stochastic gradient ascent
using the \emph{Adam} optimizer \cite{DBLP:journals/corr/KingmaB14}, a state-of-the-art stochastic optimization method with adaptive estimation of moments.

During each iteration of stochastic gradient ascent, 
we sample a user $u$, a positive item $i \in \mathcal I_u^+$, 
and a negative item  $j \in \mathcal{I}\setminus \mathcal{I}_u^+$. Thus the convolutional network $\Phi$ has \emph{two} images to consider, $\X_i$ and $\X_j$.
Following the Siamese setup \cite{hadsell2006dimensionality}, both CNNs $\Phi(\X_i)$ and $\Phi(\X_j)$ share the same weights.

\subsection{From Recommendation to Generation}

DVBPR evaluates a user preference score for a given image. 
But a richer form of recommendation might consist of guiding users and designers by helping them to explore the space of \emph{potential} fashion images and styles.
This requires a method that
can efficiently sample arbitrary realistic-looking images from a prior distribution.
Generative Adversarial Networks (GANs) \cite{goodfellow2014generative} offer an effective way to capture such a distribution and have demonstrated promising results in a variety of applications
\cite{radford2015unsupervised}. In particular, conditional GANs \cite{condgan} are a useful variant, which can generate images according to semantic inputs like class labels \cite{ACGAN}, textual descriptions \cite{reed2016generative}, etc. In our case, we train a GAN conditioned on product's top-level category. 

\begin{figure}
\centering
\includegraphics[height=5cm]{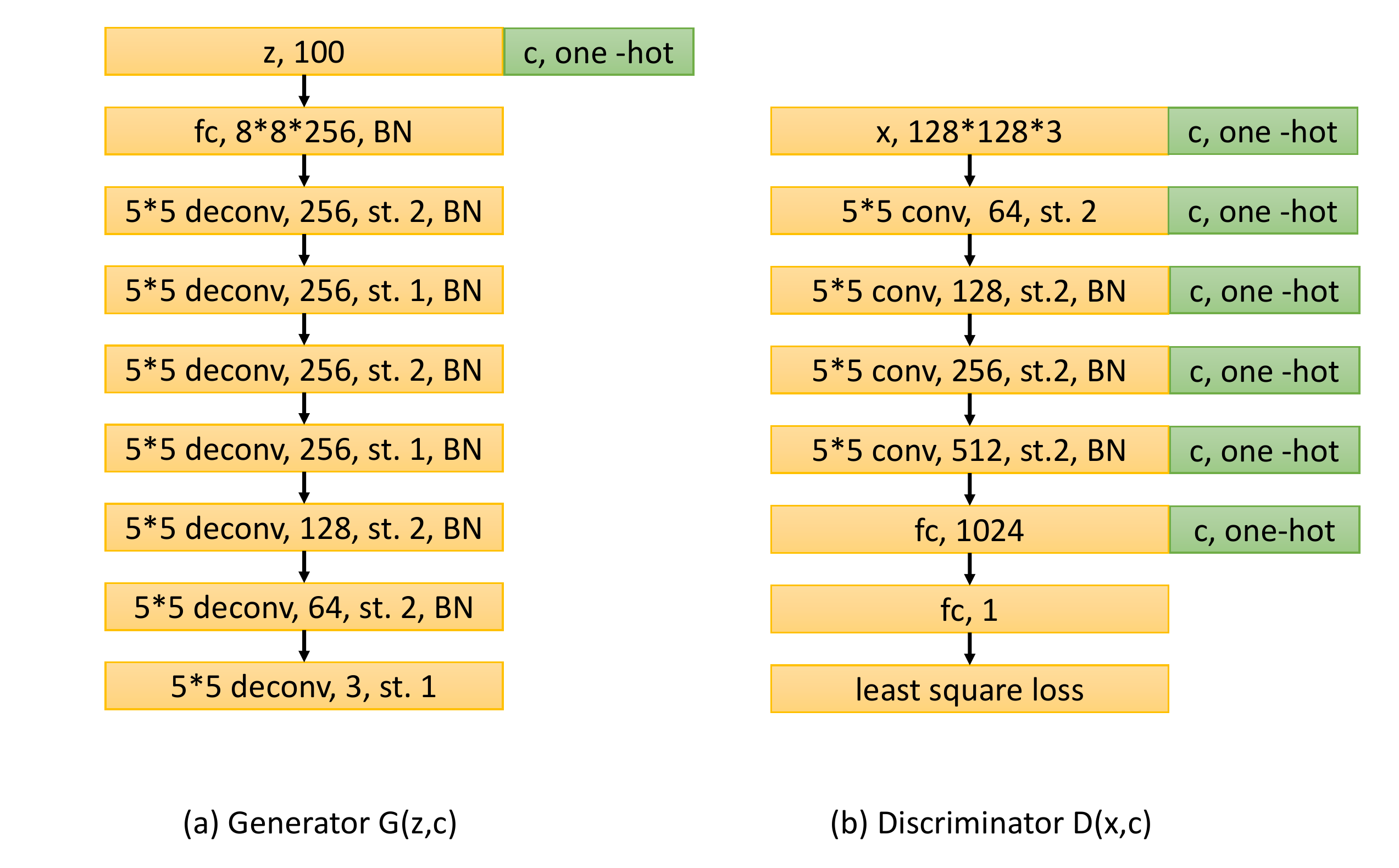}
\caption{Model architecture of our GAN models.}
\label{fig:gan_arch}
\end{figure}
A GAN consists of a generator $G$ and a discriminator $D$, which are usually implemented as multi-layer convolutional or deconvolutional neural networks.
The generator $G(\z,c)$ takes a random noise vector $\z\sim U(-\bf{1},\bf{1})$ and a category $c$ as inputs and synthesizes an image.
The discriminator takes an image $\x$ sampled either from training data $X_c$ or synthesized examples $G(\z,c)$, and predicts the likelihood of the image being `real' (i.e., belonging to the training set $X_c$). We train our GAN by using a least squares loss as in an LSGAN\cite{LSGAN}, which is a newly developed GAN with high image quality. The objective functions of our discriminator and generator are defined as:
\begin{equation}
\begin{split}
\min_D V(D)&=\mathbb{E}_{\x,c\sim p_{\text{data}}(\x,c)}L_{\mathit{real}}(\x,c)\\&+\mathbb{E}_{c\sim p(c),\z\sim p(\z)}L_{\mathit{fake}}(G(\z,c),c),\\
\min_G V(G)&=\mathbb{E}_{c\sim p(c),\z\sim p(\z)}L_{\mathit{real}}(G(\z,c),c),\label{eq:gan}\\ 
\end{split}
\end{equation}

where $L_{\mathit{real}}(\x,c)=[D(\x,c)-1]^2$ and $L_{\mathit{fake}}(\x,c)=[D(\x,c)]^2$. This means the discriminator $D$ tries to predict `1' for real images and `0' for fake images, while the generator $G$ wants to generate `realistic' images to fool $D$. These two opposing objectives are optimized alternately until the quality of generated images is acceptable (around 25 epochs in our case).
In Figure~\ref{fig:gan_arch}, we show the details of our model architectures which are deeper conditional networks modified from \cite{LSGAN,radford2015unsupervised}.  

Compared to some GAN results that generate `dream like' hallucinations of natural images, our generated images look realistic, presumably because we operate in a relatively circumscribed domain (clothing images viewed in one of a few canonical poses). This means that the samples generated by our network are distinct from, yet similar in quality to, existing images in the training corpus.

\subsection{Personalized Design}
\label{sec:personalized}

Although recommender systems are concerned with modeling users' preferences, this is typically limited to identifying items to which users would assign a high rating (or high purchase probability). Although latent variable models (and in particular matrix factorization) are able to uncover surprisingly complex semantics, they are limited in their interpretability. By using our model to generate images, we can begin to reason about what the model `knows,' and in particular explore the space of potentially desirable items that do not yet exist.


Such technology could support applications such as (1) Sampling new items, including items tailored to a specific user's preferences; (2) Tailoring existing items, i.e., making small modifications such that an item better matches the preferences of a user. These applications are useful for interpreting users' preferences learned from our model, and also for providing inspiration to designers. We describe each of these scenarios in detail in our experiments.


Our model associates preference scores between users and items (images), suggesting that by generating images that maximize this preference value we might be able to produce items that best match a user's personal style.
This process, called \emph{preference maximization}, is 
similar to activation maximization methods
\cite{DBLP:conf/nips/NguyenDYBC16},
which try to interpret deep neural networks by finding inputs that maximize a particular neuron (e.g.~a classification neuron in the final layer). 

\xhdr{Preference Maximization} Formally, given a user $u$ and a category $c$, we can straightforwardly retrieve existing items in the dataset to maximize 
a user's preference score:
\begin{equation}
\delta(u,c)=
\argmax_{e\in X_c}\ x_{u,e}
=
\argmax_{e\in X_c}\ \btheta_{u}^{T}\Phi(e),
\label{eq:delta}
\end{equation}
where $X_c$ is the set of item images belonging to category $c$.


While the above selects a `real' image 
from
the training corpus, our GAN provides us with an approximated distribution of the original data. Thus we can synthesize an image by optimizing
\begin{equation}
\begin{split}
\label{eq:delta_hat}
&\widehat{\delta}(u,c)=\argmax_{e\in G(\cdot,c)}\ \widehat{x}_{u,e}=\argmax_{e\in G(\cdot,c)}\ x_{u,e}-\eta L_{real}(e,c)\\ 
&=G\Bigg [\argmax_{\z\in \mathbb[-1,1]^{100}}\ \btheta_{u}^{T} \Phi\big[\Delta G_c(\z)\big]-\eta\big[D_c(G_c(\z))-1\big]^2, c \Bigg ],
\end{split}
\end{equation}
where $D_c(\x)=D(\x,c)$, $G_c(\z)=G(\z,c)$, $\z$ is a latent input code for the generator 
$G$
and $\Delta$ is an image upscaling operator which resizes the RGB image from $128\times 128$ to $224\times 224$. We used nearest-neighbor scaling, though any differentiable image scaling algorithm would work. We also add a term $\eta L_{\mathit{real}}(e,c)$ to control image quality via the learned discriminator. The hyper-parameter $\eta$ controls the trade-off between preference score and image quality, which we will analyze in our experiments.

To optimize \eq{eq:delta_hat} given the constraint $\z\in[-1,1]^{100}$,
we introduce an auxiliary variable $\z'\in\mathbb{R}^{100}$, and let $\z=\tanh(\z')$.
Thus our optimization problem becomes
\begin{equation}
\max_{\z'\in \mathbb{R}^{100}}\ \btheta_{u}^{T}\Phi\big[\Delta G_c(\tanh (\z'))\big]-\eta\big[D_c(G_c(\tanh (\z')))-1\big]^2,
\label{eq:ganOptAdjusted}
\end{equation}
which we solve via gradient ascent.

To sample initial points within the space we
draw $\z\sim U(-\bf{1},\bf{1})$, and 
let $\z'=\tanh^{-1}(\z)=\frac{1}{2}[\ln(1+\z)-\ln(1-\z)]$
where $\tanh^{-1}(\cdot)$ and $\ln(\cdot)$ are applied 
elementwise.
Note that such an objective can potentially have many local optima.
To get a high quality solution, we repeat the optimization process from $m$ random initial points ($m=64$ in our experiments; larger $m$ could yield better solutions but is also more time-consuming), 
choosing whichever has the highest
objective value
after optimization.

The method $\widehat{\delta}(u,c)$ can also easily adapt to generate multiple images. For example, after optimization, we rank the $m$ images $\{e_1,e_2,...,e_m\}$ according to their objective value~(i.e., $\widehat{x}_{u,e}$), and return the top-$k$ images. However, we found results from this method have poor diversity. Hence we perform sampling when returning multiple images, with the probability of choosing $e_t$ as
\begin{equation}
p(e_t)=\frac{\exp(\widehat{x}_{u,e_t})}{\sum_{d=1}^{d=m}\exp(\widehat{x}_{u,e_d})}.
\end{equation}

\xhdr{Adjusting Existing Items} 
Rather than sampling completely new items, it may be desirable to make minor alterations to those in the existing corpus, to suggest how items might be `tailored' to a user's personal style. Using generated images from GANs allows us to manipulate images through $\z$, 
which
requires us to first find an approximated image (equivalent to finding a latent code $\z$) for 
a query
image in the GAN's latent space. This can be achieved by the following optimization problem,
in terms of an $\ell_1$ reconstruction error:
\begin{equation}
\min_{\z'\in \mathbb{R}^{100}}\ \left\|G_c\left(\tanh (\z')\right)-\rDelta \X_{\text{query}}\right\|_1.
\label{eq:ganOptTweak}
\end{equation}


We consider each of the two cases above in our qualitative evaluation in the following section.


\section{Experiments}

We perform both quantitative experiments, to evaluate ranking performance and generated images,
as well as qualitative experiments to demonstrate the capacity of our system to perform image synthesis and design. (Data and code are available on the first author'’s webpage.)


\subsection{Experimental Setting}

All experiments of our method were conducted on a commodity workstation with a 4-core Intel CPU and a single GTX-1080 graphics card. Note that although our dataset contains hundreds of thousands of images, and around a million user-item interactions, it is still possible to train on commodity hardware requiring around one day of training.

\xhdr{Dataset}
Our first group of datasets were introduced in \cite{julian} and consist of reviews of clothing items crawled from \emph{Amazon.com}.
We first extract a subset
called \emph{Amazon Fashion}, which contains six representative fashion categories (men/women's tops, bottoms and shoes).
We also consider two comprehensive datasets, containing all subcategories (gloves, scarves, sunglasses, etc.), named \emph{Amazon Women} and \emph{Amazon Men}.
We treat users' reviews as implicit feedback.

Another dataset is crawled from \emph{Tradesy.com} \cite{VBPR}, which is a c2c online platform for buying and selling used fashion items.
The dataset includes several kinds of feedback, like clicks, purchases, sales, etc. We treat these as implicit feedback as in \cite{VBPR}. For all four datasets, each item 
has exactly one associated
image and a pre-extracted CNN feature  
representation.

For data preprocessing, we discard inactive users $u$ for whom $|\mathcal{I}_u^+|<5$, as in VBPR \cite{VBPR}. For each user, we randomly withhold one action for validation $\mathcal{V}_u$, and another for testing $\mathcal{T}_u$. All remaining items are used for training $\mathcal{P}_u$, and we always report performance for the model that achieved the best performance on our validation set. Statistics of our datasets are given in Table \ref{exp:dataset}. Note that we 
focus our qualitative evaluation on \emph{Amazon Fashion}, while other datasets are used to compare recommendation performance. 

\begin{table}[h]
\centering
\caption{Dataset statistics (after preprocessing) \label{exp:dataset}}
\setlength{\tabcolsep}{3pt}
\begin{tabular}{lcccc}
\toprule
Dataset & \#Users & \#Items & \#Interactions &\#Categories\\ \midrule
\emph{Amazon Fashion} & 64583   & 234892   & 513367 & 6    \\
\emph{Amazon Women}   & 97678   & 347591   & 827678 & 53     \\
\emph{Amazon Men}     & 34244   & 110636   & 254870 & 50\\
\emph{Tradesy.com}        & 33864   & 326393   & 655409 & N/A\\ \bottomrule
\end{tabular}
\setlength{\tabcolsep}{4pt}
\end{table}

\xhdr{Evaluation Metrics}
\label{sec:eva}
We calculate the
AUC to measure recommendation performance of our method and that of baselines.
The AUC 
measures the quality of a ranking based on pairwise comparisons (and is the measure that BPR-like methods are trained to optimize).
Formally, we have
\[AUC=\frac{1}{|\mathcal{U}|}\sum_{u\in\mathcal{U}} \frac{1}{|\mathcal{D}_u|}\sum_{(i,j)\in\mathcal{D}_u}\xi(x_{u,i}>x_{u,j}),\]
where $\mathcal{D}_u=\{(i,j)|(u,i)\in\mathcal{T}_u\wedge(u,j)\notin(\mathcal{P}_u\cup\mathcal{V}_u\cup\mathcal{T}_u)\}$ and $\xi(\cdot)$ is an indicator function.
In other words, we are counting the fraction of times that the `observed' items $i$ are preferred over `non-observed' items $j$.

When generating images, we use three metrics to evaluate preference scores, image quality, and diversity.
For a given user $u$, a large preference score $x_{u,i}$ suggests that the user $u$ would be interested in item $i$. 
From the perspective of our model, this ought to be true even for generated images.
Hence, the compared methods~(i.e., our retrieval-based method $\delta(u,c)$ and synthesis-based method $\widehat{\delta}(u,c)$) should find an item  with its objective value as large as possible.
Here
we randomly 
sample
a user $u$ and set $c$ to the category of the item in the test set $\mathcal{T}_u$, 
and
compute
the
mean objective value~(i.e., eq.~\ref{eq:DVBPR}) of compared methods. 
For image quality, we report the inception score\cite{IS}, which is a commonly used heuristic approach to measure image quality, based on a standard  pre-trained inception network. Higher 
scores typically mean better quality. For diversity, similar to \cite{ACGAN}, we calculate the visual similarity of pairs of returned images for each query (i.e., a user and a category), and then take the average of multiple queries. The visual similarity is measured by structural similarity (SSIM) 
\cite{SSIM}, which is more consistent with human visual perception than the mean squared error and other traditional measures. SSIM ranges from 0 to 1, where higher values indicate more similar images. We report Opposite Mean SSIM, which is one minus mean SSIM, to represent diversity. Thus a higher value means better diversity.

Both metrics for measuring image quality and diversity 
have limitations and are subjective.
Thus we also show qualitative results of synthetic images under different image generation scenarios, which we describe further in Section \ref{sec:qualitative}.


Ultimately 
while
we are unable to directly evaluate the `designed products' against real users, 
our evaluation shows that (1) Our end-to-end learning approach is state-of-the-art in terms of recommendation performance (AUC); (2) Our optimization process is able to identify plausible images that are distinct from those in the training corpus, and have higher preference scores than those of existing products.

\subsection{Baselines} When evaluating methods in terms of their AUC, we compare our method against the following baselines:

\begin{description}
\item[Random~(RAND)] Images are ranked in a random order. By definition this method has $\text{AUC}=0.5$.
\item[PopRank] Images are ranked in order of their popularity within the system.

\item[WARP] Matrix factorization for optimizing top-k performance with weighted approximated
ranking pairwise~(WARP) loss\cite{WARP}.
\item[BPR-MF] Introduced by~\cite{rendle2009bpr}, is a state-of-the-art method for personalized ranking on implicit feedback datasets. It uses standard MF (i.e., eq.~\ref{eq:standardMF}) as the underlying predictor.

Besides standard ranking methods that only use implicit feedback, we also include stronger methods that are able to exploit item visual features:

\item[VisRank] A simple content-based method based on visual similarity (i.e., distance of pre-trained CNN features) among items. Images are ranked according to their average distance to items bought by the user.

\item[Factoization Machines~(FM)] Proposed by \cite{FM}, provides a generic factorization approach that can be used to estimate the score of an item given by a user. We use the same BPR loss to optimize personalized ranking. Also as in \cite{lightFM}, we only consider interactions between (user, item) and (user, item's CNN feature).

\item[VBPR] A state-of-the-art method for visually-aware personalized ranking from implicit feedback \cite{VBPR}. VBPR is a form of hybrid content-aware recommendation that makes use of
pre-trained CNN features of product images. 

\item[DVBPR] The method proposed in this paper.
\end{description}

These baselines are intended to show (a) the importance of learning personalized notions of compatibility for fashion recommendation (MF methods vs.~PopRank); (b) the improvement to be gained by incorporating visual features directly into the recommendation pipeline (FM/VBPR/DVBPR vs.~MF methods); and (c) the improvement to be gained by learning image representations expressly for the task of fashion recommendation, rather than relying on features from a pre-trained model (DVBPR vs.~FM/VBPR). Note that there are alternative methods that make use of other side information (e.g. social or temporal) for recommendation \cite{HeFanWanMcA,upsdowns}, though these are orthogonal to our approach. We also considered related methods based on the Siamese setup \cite{SiameseICCV,lei2016comparative}, though these were not directly comparable as they optimize different objectives or consider different problem settings (item-to-item recommendation, user features, etc.).

For WARP and FM, we use LightFM's\cite{lightFM} implementation; for other baselines we use our own implementation. DVBPR is implemented in \emph{TensorFlow}.

For WARP, BPR-MF, FM and VBPR, we tune hyper-parameters via grid search on a validation set, with a regularizer selected from $\{0,0.001,0.01,0.1,1,10,30\}$ and the number of latent factors selected from $\{10,30,50\}$. We found that DVBPR is not overly sensitive to hyper-parameter tuning, hence we use a single setting for all datasets. We used a mini-batch size of 128 for DVBPR and 64 for GAN training. We set the number of latent factors $K=50$ and regularization hyper-parameter $\lambda_{\theta_u}=1$. Regularization of item factors is achieved via weight decay and dropout of the CNN-F network.


\begin{table*}
\centering
\caption{Recommendation performance in terms of the AUC (higher is better). The best performing method in each row is boldfaced.
\label{tab:recommendation}}
\begin{tabular}{llcccccccccc}
\toprule
\multirow{2}{*}{Dataset} & \multirow{2}{*}{Setting}     & \multirow{2}{*}{\begin{tabular}[c]{@{}c@{}}(a)\\ RAND\end{tabular}} & \multirow{2}{*}{\begin{tabular}[c]{@{}c@{}}(b)\\ PopRank\end{tabular}} & \multirow{2}{*}{\begin{tabular}[c]{@{}c@{}}(c)\\  WARP\end{tabular}} & \multirow{2}{*}{\begin{tabular}[c]{@{}c@{}}(d)\\ BPR-MF\end{tabular}} & \multirow{2}{*}{\begin{tabular}[c]{@{}c@{}}(e)\\ VisRank\end{tabular}} &\multirow{2}{*}{\begin{tabular}[c]{@{}c@{}}(f)\\ FM\end{tabular}}& \multirow{2}{*}{\begin{tabular}[c]{@{}c@{}}(g)\\ VBPR\end{tabular}} & \multirow{2}{*}{\begin{tabular}[c]{@{}c@{}}(h)\\ DVBPR\end{tabular}} & \multicolumn{2}{c}{Improvement}\\
 & & & & & & & & & &\multicolumn{1}{c}{h vs.~d} & \multicolumn{1}{c}{h vs.~best} \\ \midrule
\multirow{2}{*}{\emph{Amazon Fashion}} & All Items  & 0.5 & 0.5849 & 0.6065 & 0.6278 & 0.6839 & 0.7093 & 0.7479 & \textbf{0.7964} & 26.9\% & 6.5\% \\
                                       & Cold Items & 0.5 & 0.3905 & 0.5030 & 0.5514 & 0.6807 & 0.7088 & 0.7319 & \textbf{0.7718} & 40.0\% & 5.5\% \\[1.5mm]
\multirow{2}{*}{\emph{Amazon Women}}    & All Items  & 0.5 & 0.6192 & 0.5998 & 0.6543 & 0.6512 & 0.6678 & 0.7081 & \textbf{0.7574} & 15.8\% & 7.0\% \\
                                        & Cold Items & 0.5 & 0.3822 & 0.5017 & 0.5196 & 0.6387 & 0.6682 & 0.6885 & \textbf{0.7137} & 37.4\% & 3.7\% \\[1.5mm]
\multirow{2}{*}{\emph{Amazon Men}}      & All Items  & 0.5 & 0.6060 & 0.6081 & 0.6450 & 0.6589 & 0.6654 & 0.7089 & \textbf{0.7410} & 14.9\% & 4.5\% \\
                                        & Cold Items & 0.5 & 0.3903 & 0.5005 & 0.5132 & 0.6545 & 0.6705 & 0.6863 & \textbf{0.6923} & 34.9\% & 0.9\% \\[1.5mm]
\multirow{2}{*}{\emph{Tradesy.com}}      & All Items  & 0.5 & 0.5268 & 0.6176  & 0.5860 & 0.6457 & 0.7662 & 0.7500 & \textbf{0.7857} & 34.1\% & 2.5\% \\
                                        & Cold Items & 0.5 & 0.3946 & 0.5333 & 0.5418 & 0.6084 & 0.7730 & 0.7525 & \textbf{0.7793} & 43.8\% & 0.8\% \\                                        
\bottomrule
\end{tabular}
\end{table*}

\subsection{Quantitative Evaluation}

\xhdr{Recommendation Performance}
We report recommendation performance in terms of the AUC in Table \ref{tab:recommendation}. We consider two settings, \emph{All Items} and \emph{Cold Items}. For the latter, we seek to estimate relative preference scores among items that have few observations at training time (fewer than 5). 


On average, our method outperforms the second-best method by 5.13\% across all datasets, and 2.73\% in cold-start scenarios. Our method outperforms the strongest content-\emph{un}aware method (BPR-MF) substantially; in fact, even our simple `nearest-neighbor' style baseline (VisRank) outperforms content-unaware methods on this data. The poor performance of MF based methods is largely due to the extreme sparsity of the data (2-3 observations per item on average) and is consistent with previous reported results on the same datasets \cite{VBPR}.
This confirms the importance of content-aware recommendation methods on extremely long-tailed applications like fashion recommendation.

\xhdr{Evaluation of Generated Images}

Next, our goal is to quantitatively compare real versus generated images. As we stated in Section \ref{sec:eva}, we use the mean objective value (i.e., eq.~\ref{eq:DVBPR}), inception score \cite{IS} and Opposite SSIM \cite{ACGAN,SSIM} to evaluate how well images match user preferences, image quality, and diversity, respectively.

Table \ref{tb:comparison_real_fake} shows evaluation results of images from the four sources. The first two methods are random methods, which draw real images (from $X_c$), or generated images (from $G(\cdot,c)$), by feeding uniform random vectors $\z$. 
Both methods exhibit poor performance in term of preference score, since they don't consider the given user in each query. However, the generated images can achieve comparable performance regarding quality and diversity, which verifies the effectiveness of GAN training in that the GAN can generate realistic and diverse (rather than noisy or identical) images. 

The last two methods $\delta(u,c)$ and $\widehat{\delta}(u,c)$ are personalized methods, hence they both achieve higher preference scores compared to random methods. In particular, even in a comprehensive dataset like \emph{Amazon's} clothing catalog, a synthesis-based method ($\widehat{\delta}(u,c)$) can gain 6.8\% improvement over a retrieval-based method ($\delta(u,c)$) in terms of preference score. This shows the generative model can uncover potentially desirable items
that don't exist in the dataset.
Furthermore, $\widehat{\delta}(u,c)$ achieves similar quality
and slightly lower diversity compared to $\delta(u,c)$.

\begin{table}[t]
\centering
\caption{Comparison of top-3 returned images for a given user and category (1000 trials). For all three metrics, a larger value is better. The value after the $\pm$ is the standard deviation.}
\label{tb:comparison_real_fake}
\begin{tabular}{lccc}
\toprule
Item Source             & Preference Score       & Quality       &Diversity      \\ \midrule
Random Methods&&&\\
$\hspace{1em}X_c$  & -1.9547$\pm$3.8 & 6.8685$\pm$.36 & 0.5585$\pm$.09 \\
$\hspace{1em}G(\cdot,c)$     &  -1.9392$\pm$3.7 & 6.8121$\pm$.37 & 0.5589$\pm$.09                     \\
Personalized Methods&&&\\
$\hspace{1em}\delta(u,c)$ in (eq.~\ref{eq:delta})        & 7.1876$\pm$3.2  & 7.6492$\pm$.27 & 0.5393$\pm$.11            \\
$\hspace{1em}\widehat{\delta}(u,c)$ in (eq.~\ref{eq:delta_hat})         & 7.6760$\pm$4.1  & 7.6524$\pm$.24 & 0.5265$\pm$.12
      \\ \bottomrule
\end{tabular}
\end{table}

\begin{figure*}
\centering
\subfigure[Preference Score]{\raisebox{-9pt}{\includegraphics[width=0.3\linewidth]{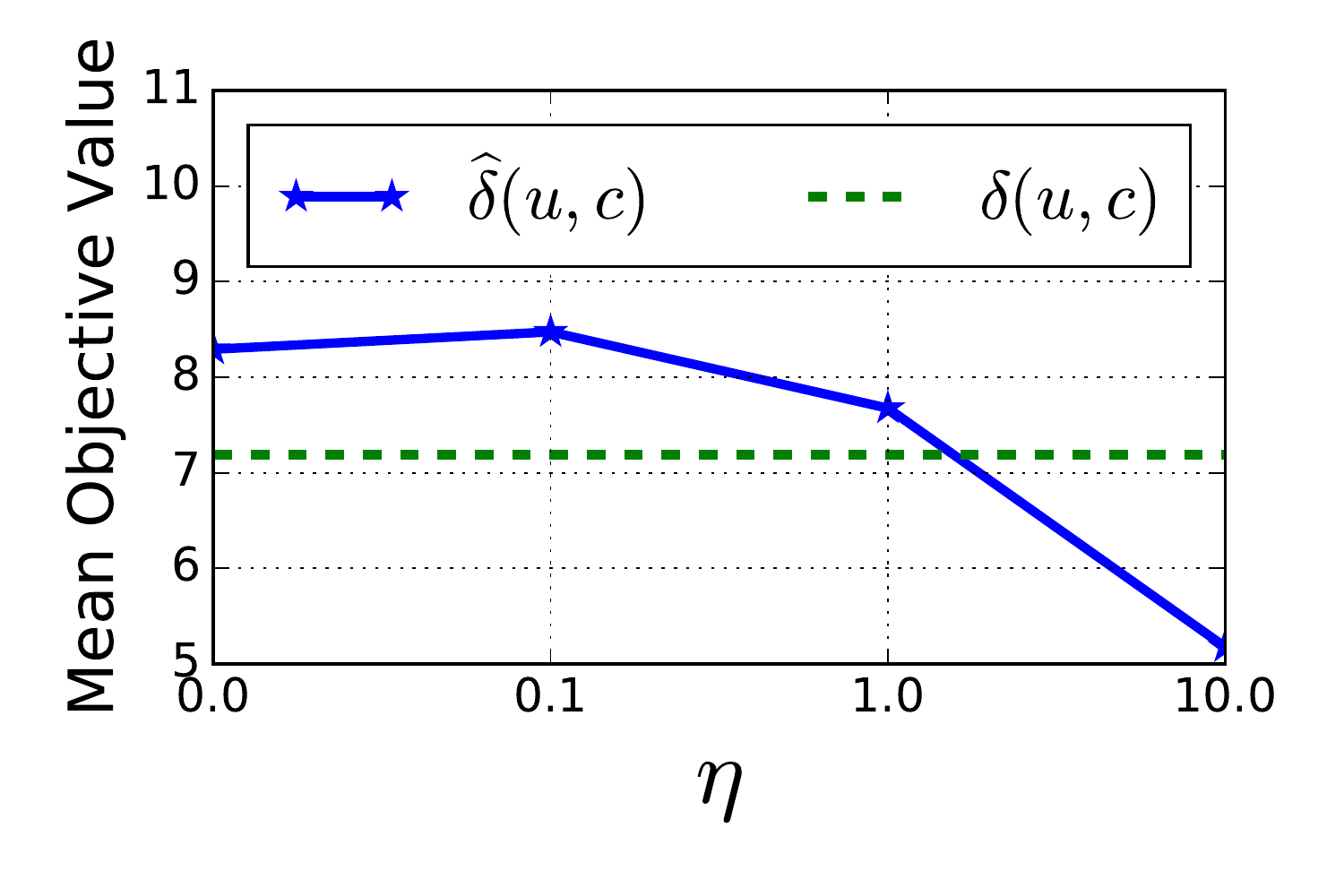}}}
\subfigure[Image Quality]{\raisebox{-9pt}{\includegraphics[width=0.3\linewidth]{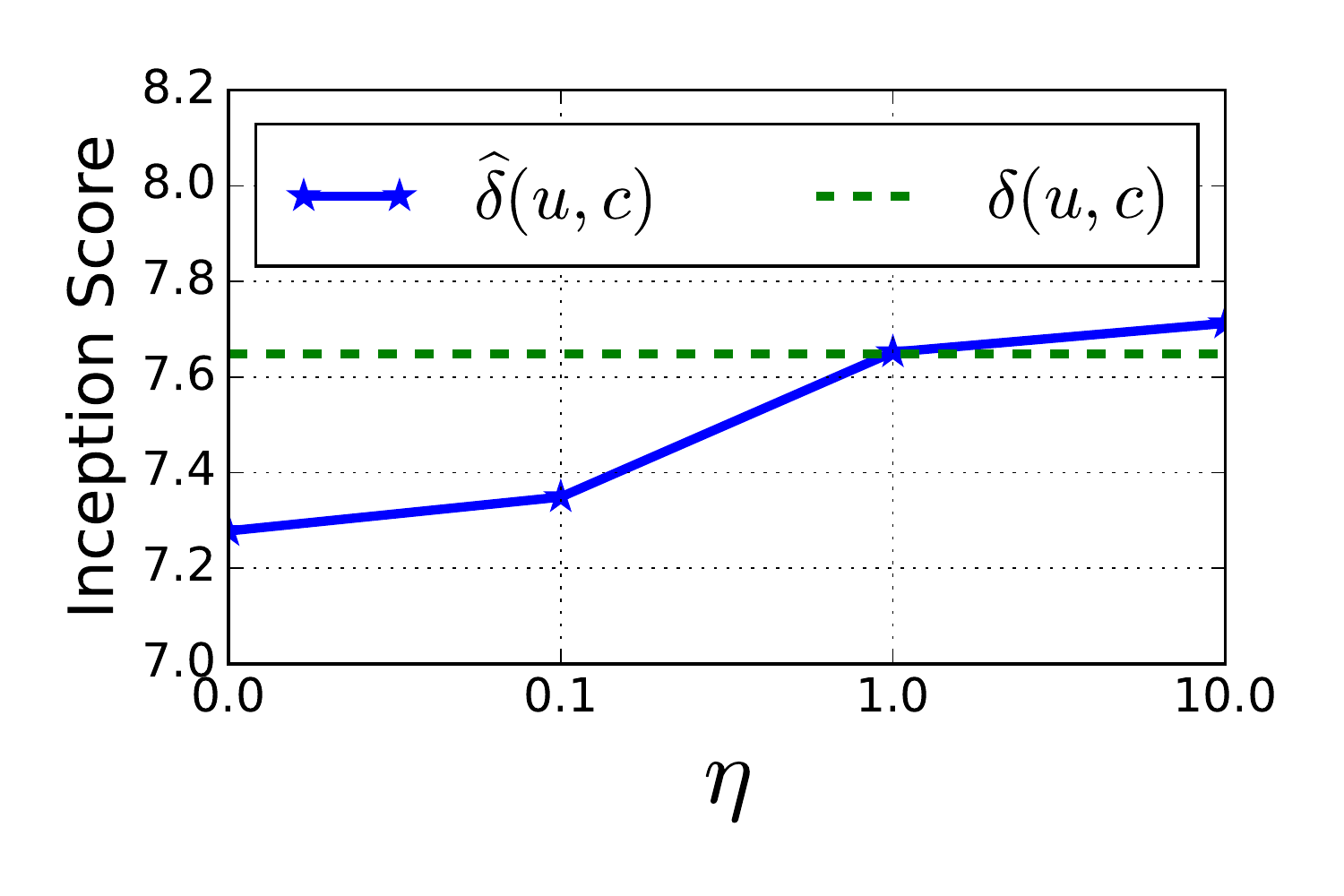}}}
\subfigure[Image Diversity]{\raisebox{-9pt}{\includegraphics[width=0.3\linewidth]{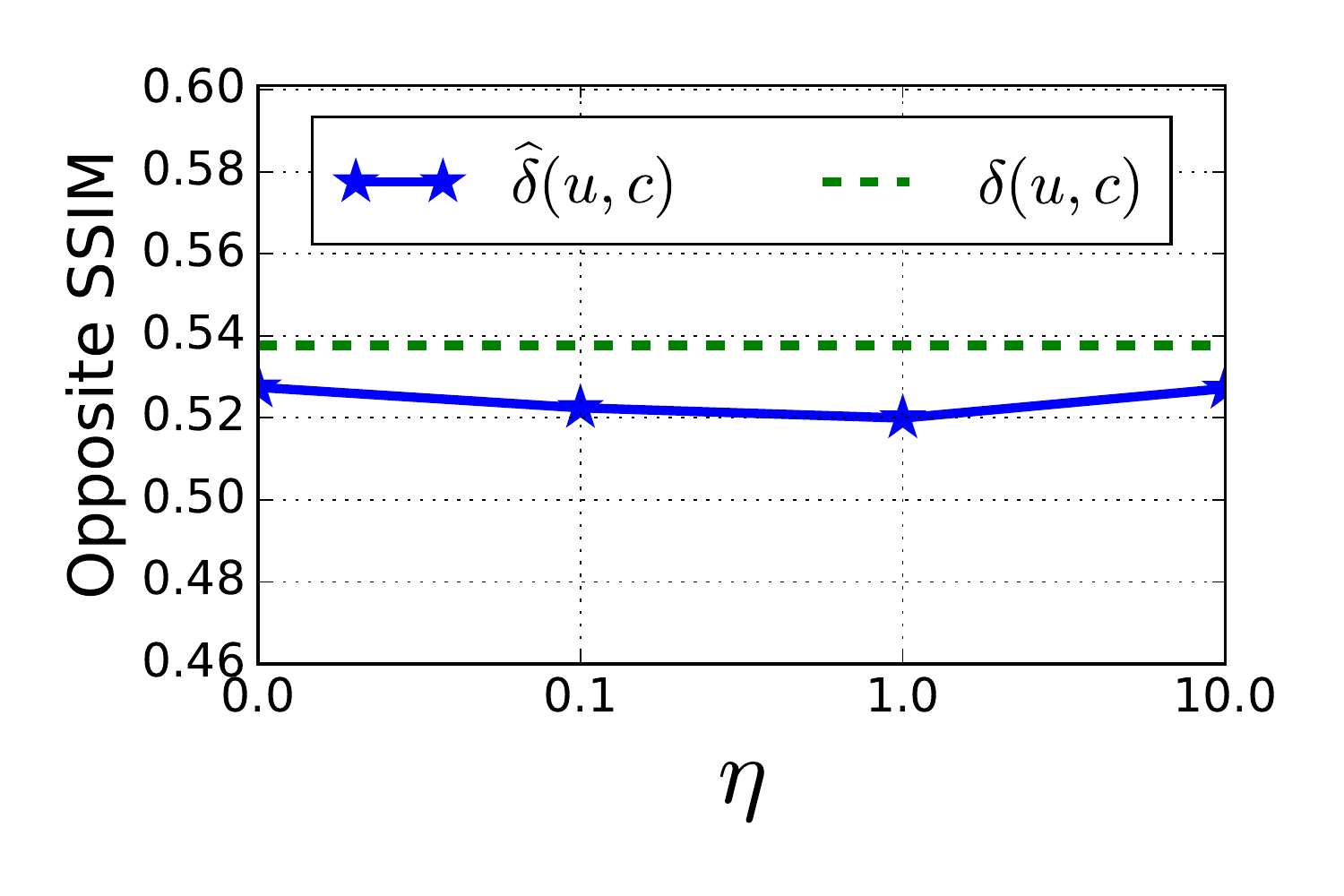}}}
\caption{Effect of hyper-parameter $\eta$.}
\label{fig:eta}
\end{figure*}

In the method $\widehat{\delta}(u,c)$, there is an important hyper-parameter $\eta$ to control the trade-off between preference score and quality. In Figure \ref{fig:eta}, we plot the performance of the three metrics under different $\eta$. It is clear that as $\eta$ increases, the preference score drops and the quality becomes better, which is consistent with our intention. However $\eta$ doesn't have an obvious impact on diversity. According to these results, we choose $\eta=1$ as the `sweet spot' in all experiments, as it results in better preference scores and similar quality compared to $\delta(u,c)$.

\subsection{Qualitative Evaluation}
\label{sec:qualitative}

The above experiments suggest a form of recommender system that can aid in the design and exploration of new items. Finally, we assess our system qualitatively, by demonstrating image generation results following each of the settings in Section \ref{sec:personalized}: image sampling, personalized design, and prototype-based tailoring.

\begin{figure}
\centering
\includegraphics[width=\linewidth]{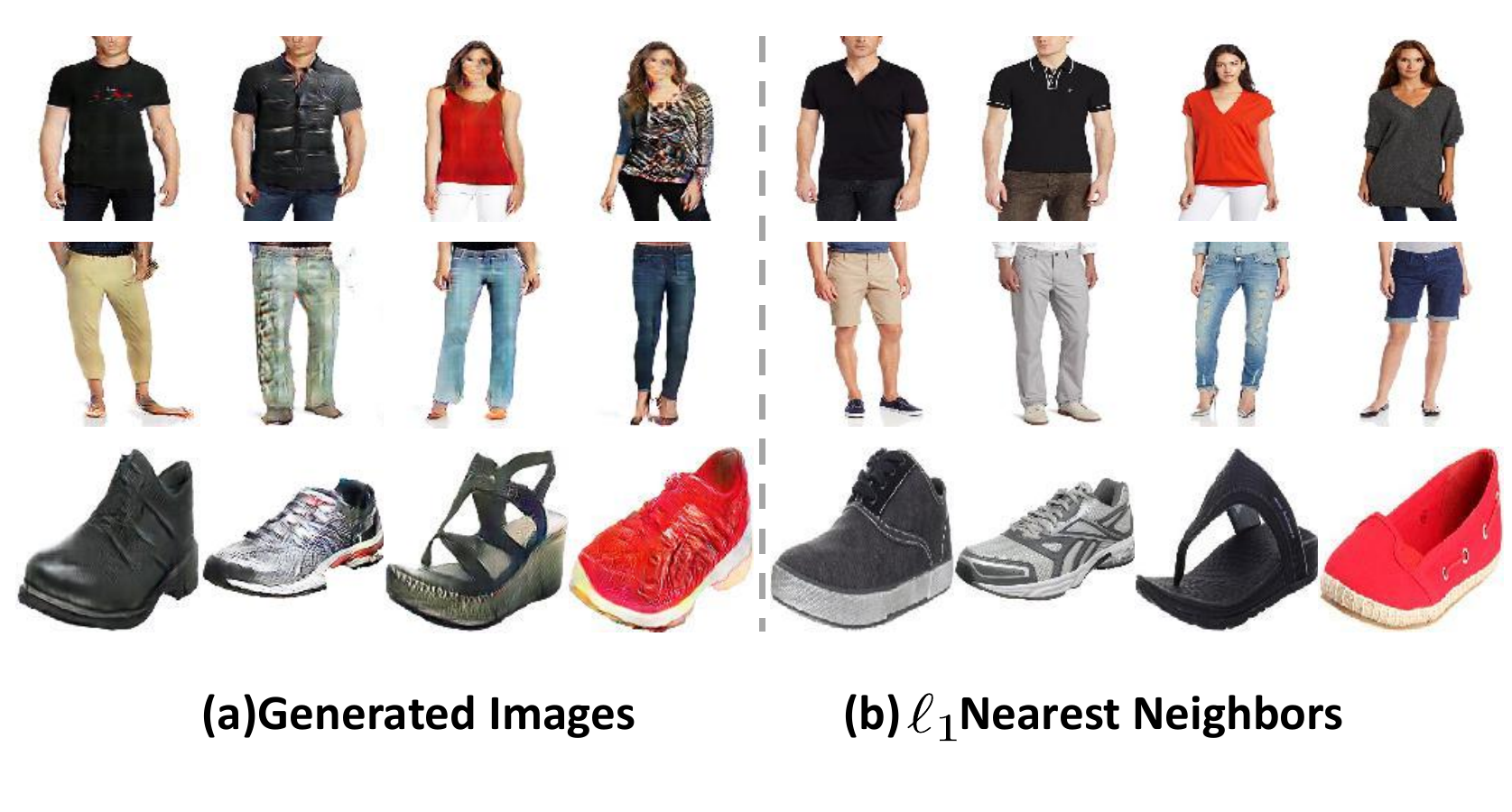}
 \vspace{-1cm}
\caption{Generated image samples and their $\ell_1$ nearest neighbors in the dataset ($\ell_1$ and $\ell_2$ nearest neighbors were equivalent in almost all cases). Note that all images are rescaled to be square during preprocessing.
\label{fig:l1NN}}
\end{figure}

First, Figure \ref{fig:l1NN} compares sampled images from each of six categories (women/men's tops, bottoms and shoes) to their nearest neighbors in the dataset. This demonstrates that the generated images are realistic and plausible, yet are quite different from any images in the original dataset---they have common shape and color profiles, but quite different styles.


\xhdr{Fashion design for a given user}
Next, we examine images that have been personalized to a single user (i.e., Top-3 returned images from $\delta(u,c)$ and $\widehat{\delta}(u,c)$).

\begin{figure}[t]
\centering
\includegraphics[width=\linewidth]{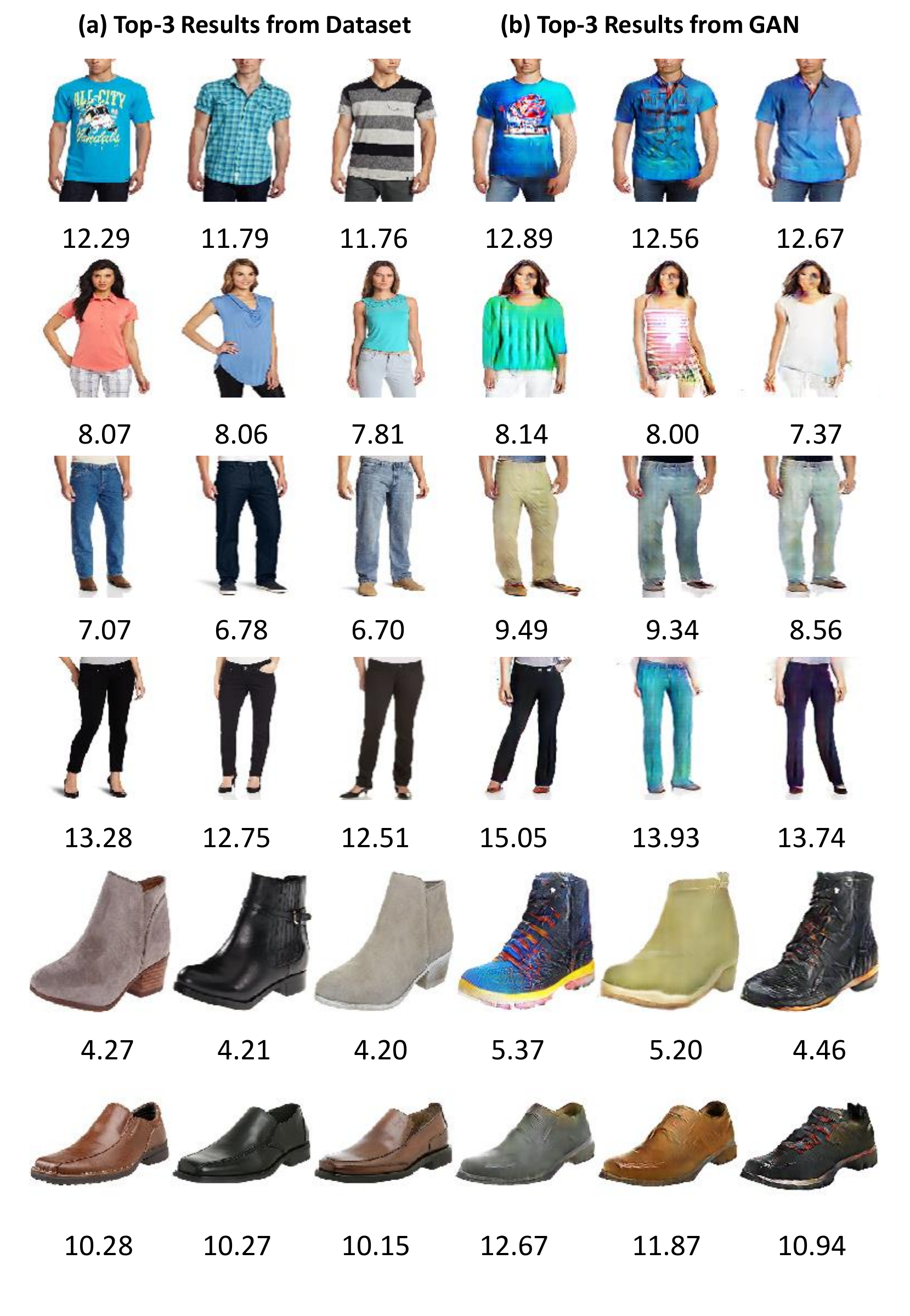}

\vspace{-0.7cm}
\caption{Top-3 results from the dataset and GAN. Each row is a separate retrieval/optimization process for a given user and product category. 
Left: real images; right: synthetic images.
The values shown are preference scores for each image.
\label{fig:personalized}
}
\vspace{-3mm}
\end{figure}

Results for six users are shown in Figure \ref{fig:personalized}. We select the top 3 images generated by the GAN (in terms of personalized objective value) with the top 3 images in the dataset. Note that while the images generated are again quite different from those in the training corpus, they seem to belong to a similar style, indicating that the users' individual styles have been effectively captured.

\xhdr{Fashion design based on a prototype}
The second type of personalized design consists of making small modifications to existing items so that they 
more closely match users' preferences.
 
\begin{figure}[t]
\centering
\includegraphics[width=0.96\linewidth]{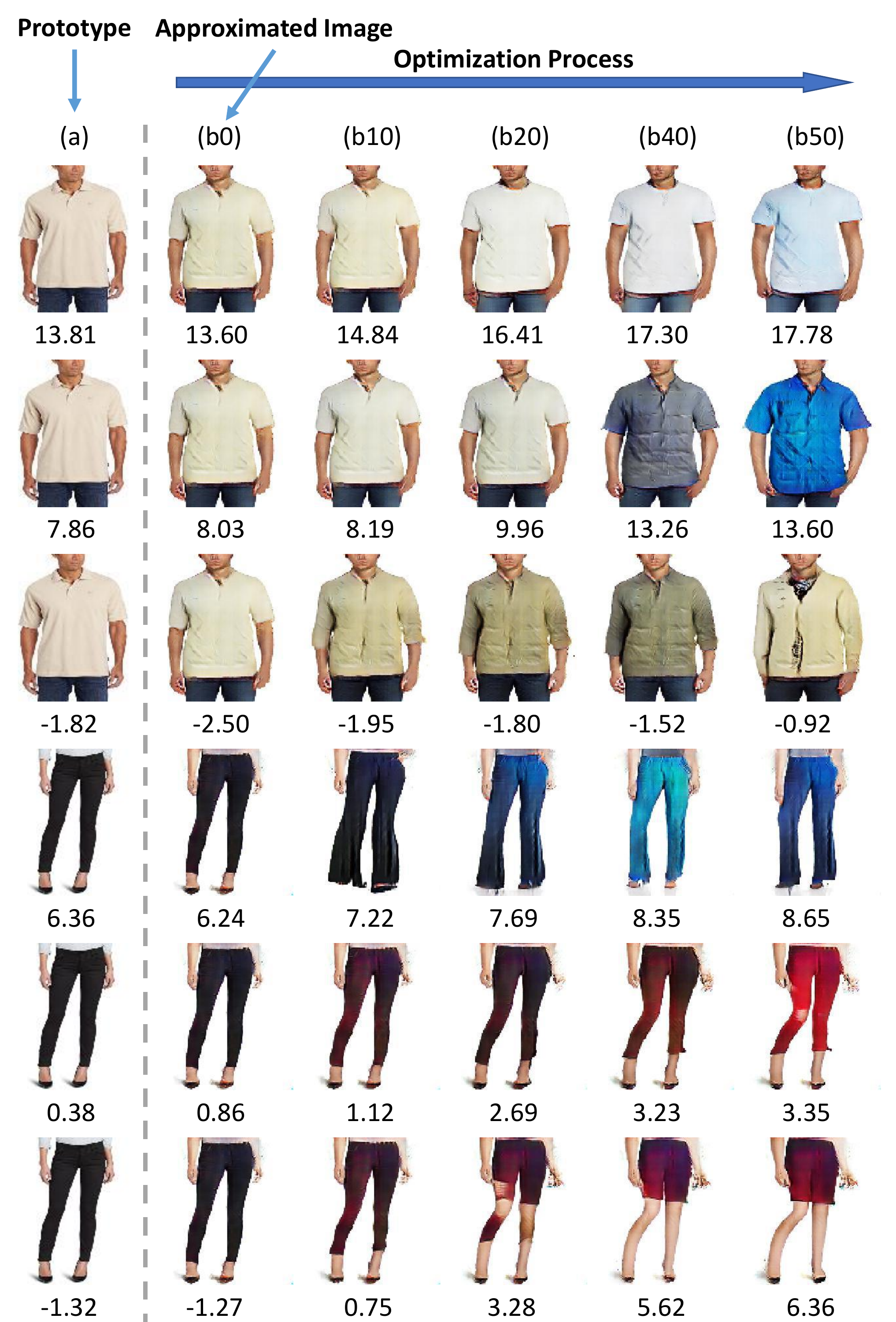}

\vspace{-3mm}
\caption{Optimization for different users given the same initial image. Each row is a separate process for a different user, given the real image in column (a).
Columns (b0) through (b50) show the image after successive iterations of GAN optimization.
The preference score of each image is shown below the image.
\label{fig:localOpt}}
\vspace{-5mm}
\end{figure}

Figure \ref{fig:localOpt} shows the results of this optimization process (described in eqs.~\ref{eq:ganOptTweak} and \ref{eq:ganOptAdjusted}) for six users based on a given t-shirt or pair of pants.
There are a few relevant observations to make: First, the optimization process substantially increases the preference score over that of the original image. Second, the L1-approximated images are similar to the given prototype, and most images are in the vicinity of the original image. Third, the process highlights the continuous nature of the space learned by the GAN. The types of modifications vary for different users, including but not limited to changing color, extending sleeve length (row 3), `distressing' pants (row 5), shortening pants (row 6) and other minor stylistic changes.

\section{Conclusions and Future Work}

We have presented a system for fashion recommendation that is capable not only of suggesting existing items to a user, but which is also capable of generating new, plausible fashion images that match user preferences. This suggests a new type of recommendation approach that can be used for both prediction and design.

We extended previous work on visually-aware recommendation using an end-to-end learning approach based on the Siamese-CNN framework. This leads to substantially more accurate performance than methods that make use of pre-trained representations. We then used the framework of Generative Adversarial Networks to learn the distribution of fashion images and generate novel fashion items that maximize users' preferences. This framework can be used in a variety of recommendation scenarios, to explore the space of possible fashion items, to modify existing items, or to generate items tailored to an individual.

In the future, we believe this opens up a promising line of work in using recommender systems for design. Other than improving the quality of the generated images and providing control of fine-grained styles, the same ideas can be applied to visual data besides fashion images, or even to non-visual forms of content. We believe that such frameworks can lead to richer forms of recommendation, where content recommendation and content generation are more closely linked.




\scriptsize
\bibliographystyle{IEEEtran}
\bibliography{sigproc}

\end{document}

%% file: defs.tex
\def\f{{\bf f}}

\def\x{{\bf x}}

\def\z{{\bf z}}

\def\E{{\bf E}}

\def\X{{\bf X}}

\def\0{{\bf 0}}
\def\1{{\bf 1}}
\def\2{{\bf 2}}
\def\3{{\bf 3}}
\def\4{{\bf 4}}
\def\5{{\bf 5}}
\def\6{{\bf 6}}
\def\7{{\bf 7}}
\def\8{{\bf 8}}
\def\9{{\bf 9}}

\def\bgamma{{\bm{\gamma}}}
\def\btheta{{\bm{\theta}}}
\def\rDelta{{\rotatebox[origin=c]{180}{$\Delta$}}}

%% file: bare_conf.bbl
\begin{thebibliography}{10}
\providecommand{\url}[1]{#1}
\csname url@samestyle\endcsname
\providecommand{\newblock}{\relax}
\providecommand{\bibinfo}[2]{#2}
\providecommand{\BIBentrySTDinterwordspacing}{\spaceskip=0pt\relax}
\providecommand{\BIBentryALTinterwordstretchfactor}{4}
\providecommand{\BIBentryALTinterwordspacing}{\spaceskip=\fontdimen2\font plus
\BIBentryALTinterwordstretchfactor\fontdimen3\font minus
  \fontdimen4\font\relax}
\providecommand{\BIBforeignlanguage}[2]{{%
\expandafter\ifx\csname l@#1\endcsname\relax
\typeout{** WARNING: IEEEtran.bst: No hyphenation pattern has been}%
\typeout{** loaded for the language `#1'. Using the pattern for}%
\typeout{** the default language instead.}%
\else
\language=\csname l@#1\endcsname
\fi
#2}}
\providecommand{\BIBdecl}{\relax}
\BIBdecl

\bibitem{VBPR}
R.~He and J.~McAuley, ``{VBPR:} visual bayesian personalized ranking from
  implicit feedback,'' in \emph{AAAI}, 2016.

\bibitem{CAFFE}
Y.~Jia, E.~Shelhamer, J.~Donahue, S.~Karayev, J.~Long, R.~B. Girshick,
  S.~Guadarrama, and T.~Darrell, ``Caffe: Convolutional architecture for fast
  feature embedding,'' in \emph{MM}, 2014.

\bibitem{upsdowns}
R.~He and J.~McAuley, ``Ups and downs: Modeling the visual evolution of fashion
  trends with one-class collaborative filtering,'' in \emph{WWW}, 2016.

\bibitem{HeFanWanMcA}
R.~He, C.~Fang, Z.~Wang, and J.~McAuley, ``Vista: A visually, socially, and
  temporally-aware model for artistic recommendation,'' in \emph{RecSys}, 2016.

\bibitem{VeiKovBelMcABalBel15}
A.~Veit, B.~Kovacs, S.~Bell, J.~McAuley, K.~Bala, and S.~Belongie, ``Learning
  visual clothing style with heterogeneous dyadic co-occurrences,'' in
  \emph{ICCV}, 2015.

\bibitem{wang2015collaborative}
H.~Wang, N.~Wang, and D.-Y. Yeung, ``Collaborative deep learning for
  recommender systems,'' in \emph{SIGKDD}, 2015.

\bibitem{lei2016comparative}
C.~Lei, D.~Liu, W.~Li, Z.-J. Zha, and H.~Li, ``Comparative deep learning of
  hybrid representations for image recommendations,'' in \emph{CVPR}, 2016.

\bibitem{rendle2009bpr}
S.~Rendle, C.~Freudenthaler, Z.~Gantner, and L.~Schmidt-Thieme, ``{BPR:}
  bayesian personalized ranking from implicit feedback,'' in \emph{UAI}, 2009.

\bibitem{goodfellow2014generative}
I.~Goodfellow, J.~Pouget-Abadie, M.~Mirza, B.~Xu, D.~Warde-Farley, S.~Ozair,
  A.~Courville, and Y.~Bengio, ``Generative adversarial nets,'' in \emph{NIPS},
  2014.

\bibitem{hadsell2006dimensionality}
R.~Hadsell, S.~Chopra, and Y.~LeCun, ``Dimensionality reduction by learning an
  invariant mapping,'' in \emph{CVPR}, 2006.

\bibitem{DBLP:conf/nips/NguyenDYBC16}
A.~Nguyen, A.~Dosovitskiy, J.~Yosinski, T.~Brox, and J.~Clune, ``Synthesizing
  the preferred inputs for neurons in neural networks via deep generator
  networks,'' in \emph{NIPS}, 2016.

\bibitem{julian}
J.~McAuley, C.~Targett, Q.~Shi, and A.~van~den Hengel, ``Image-based
  recommendations on style and substitutes,'' in \emph{SIGIR}, 2015.

\bibitem{WRMF}
Y.~Hu, Y.~Koren, and C.~Volinsky, ``Collaborative filtering for implicit
  feedback datasets,'' in \emph{ICDM}, 2008.

\bibitem{OCCF}
R.~Pan, Y.~Zhou, B.~Cao, N.~N. Liu, R.~Lukose, M.~Scholz, and Q.~Yang,
  ``One-class collaborative filtering,'' in \emph{ICDM}, 2008.

\bibitem{DBLP:conf/icdm/YangLTXLZ15}
J.~Yang, C.~Liu, M.~Teng, H.~Xiong, M.~Liao, and V.~Zhu, ``Exploiting temporal
  and social factors for {B2B} marketing campaign recommendations,'' in
  \emph{ICDM}, 2015.

\bibitem{DBLP:conf/icdm/LianGZYXZR15}
D.~Lian, Y.~Ge, F.~Zhang, N.~J. Yuan, X.~Xie, T.~Zhou, and Y.~Rui,
  ``Content-aware collaborative filtering for location recommendation based on
  human mobility data,'' in \emph{ICDM}, 2015.

\bibitem{DBLP:conf/icdm/YaoFLLX16}
Z.~Yao, Y.~Fu, B.~Liu, Y.~Liu, and H.~Xiong, ``{POI} recommendation: {A}
  temporal matching between {POI} popularity and user regularity,'' in
  \emph{ICDM}, 2016.

\bibitem{covington2016deep}
P.~Covington, J.~Adams, and E.~Sargin, ``Deep neural networks for youtube
  recommendations,'' in \emph{RecSys}, 2016.

\bibitem{DBLP:conf/www/WangWTSRL17}
S.~Wang, Y.~Wang, J.~Tang, K.~Shu, S.~Ranganath, and H.~Liu, ``What your images
  reveal: Exploiting visual contents for point-of-interest recommendation,'' in
  \emph{WWW}, 2017.

\bibitem{DBLP:journals/tmm/LiCZL17}
Y.~Li, L.~Cao, J.~Zhu, and J.~Luo, ``Mining fashion outfit composition using an
  end-to-end deep learning approach on set data,'' \emph{TMM}, 2017.

\bibitem{DBLP:conf/mm/HanWJD17}
X.~Han, Z.~Wu, Y.~Jiang, and L.~S. Davis, ``Learning fashion compatibility with
  bidirectional lstms,'' in \emph{MM}, 2017.

\bibitem{DBLP:conf/iccv/Al-HalahSG17}
Z.~Al{-}Halah, R.~Stiefelhagen, and K.~Grauman, ``Fashion forward: Forecasting
  visual style in fashion,'' in \emph{ICCV}, 2017.

\bibitem{bossard2013apparel}
L.~Bossard, M.~Dantone, C.~Leistner, C.~Wengert, T.~Quack, and L.~Van~Gool,
  ``Apparel classification with style,'' in \emph{ACCV}, 2013.

\bibitem{murillo2012urban}
A.~C. Murillo, I.~S. Kwak, L.~Bourdev, D.~Kriegman, and S.~Belongie, ``Urban
  tribes: Analyzing group photos from a social perspective,'' in \emph{Computer
  Vision and Pattern Recognition Workshops (CVPRW)}, 2012.

\bibitem{runway2realwayWACV15}
S.~Vittayakorn, K.~Yamaguchi, A.~C. Berg, and T.~L. Berg, ``Runway to realway:
  Visual analysis of fashion,'' in \emph{WACV}, 2015.

\bibitem{krizhevsky2012imagenet}
A.~Krizhevsky, I.~Sutskever, and G.~E. Hinton, ``Imagenet classification with
  deep convolutional neural networks,'' in \emph{NIPS}, 2012.

\bibitem{hu2014discriminative}
J.~Hu, J.~Lu, and Y.-P. Tan, ``Discriminative deep metric learning for face
  verification in the wild,'' in \emph{CVPR}, 2014.

\bibitem{bell15siggraph}
S.~Bell and K.~Bala, ``Learning visual similarity for product design with
  convolutional neural networks,'' in \emph{TOG}, 2015.

\bibitem{SiameseICCV}
A.~Veit, B.~Kovacs, S.~Bell, J.~McAuley, K.~Bala, and S.~Belongie, ``Learning
  visual clothing style with heterogeneous dyadic co-occurrences,'' in
  \emph{ICCV}, 2015.

\bibitem{condgan}
M.~Mirza and S.~Osindero, ``Conditional generative adversarial nets,''
  \emph{arXiv preprint arXiv:1411.1784}, 2014.

\bibitem{korenSurvey}
Y.~Koren and R.~Bell, ``Advances in collaborative filtering,'' in
  \emph{Recommender Systems Handbook}.\hskip 1em plus 0.5em minus 0.4em\relax
  Springer, 2011.

\bibitem{chatfield2014return}
K.~Chatfield, K.~Simonyan, A.~Vedaldi, and A.~Zisserman, ``Return of the devil
  in the details: Delving deep into convolutional nets,'' in \emph{BMVC}, 2014.

\bibitem{he2016deep}
K.~He, X.~Zhang, S.~Ren, and J.~Sun, ``Deep residual learning for image
  recognition,'' in \emph{CVPR}, 2016.

\bibitem{DBLP:journals/corr/KingmaB14}
D.~P. Kingma and J.~Ba, ``Adam: {A} method for stochastic optimization,'' in
  \emph{ICLR}, 2015.

\bibitem{radford2015unsupervised}
A.~Radford, L.~Metz, and S.~Chintala, ``Unsupervised representation learning
  with deep convolutional generative adversarial networks,'' \emph{arXiv
  preprint arXiv:1511.06434}, 2015.

\bibitem{ACGAN}
A.~Odena, C.~Olah, and J.~Shlens, ``Conditional image synthesis with auxiliary
  classifier gans,'' in \emph{ICML}, 2017.

\bibitem{reed2016generative}
S.~Reed, Z.~Akata, X.~Yan, L.~Logeswaran, B.~Schiele, and H.~Lee, ``Generative
  adversarial text to image synthesis,'' in \emph{ICML}, 2016.

\bibitem{LSGAN}
X.~Mao, Q.~Li, H.~Xie, R.~Y. Lau, Z.~Wang, and S.~P. Smolley, ``Least squares
  generative adversarial networks,'' \emph{arXiv preprint ArXiv:1611.04076},
  2016.

\bibitem{IS}
T.~Salimans, I.~J. Goodfellow, W.~Zaremba, V.~Cheung, A.~Radford, and X.~Chen,
  ``Improved techniques for training gans,'' in \emph{NIPS}, 2016.

\bibitem{SSIM}
Z.~Wang, A.~C. Bovik, H.~R. Sheikh, and E.~P. Simoncelli, ``Image quality
  assessment: from error visibility to structural similarity,'' \emph{TIP},
  2004.

\bibitem{WARP}
J.~Weston, S.~Bengio, and N.~Usunier, ``Wsabie: Scaling up to large vocabulary
  image annotation,'' in \emph{IJCAI}, 2011.

\bibitem{FM}
S.~Rendle, ``Factorization machines,'' in \emph{ICDM}, 2010.

\bibitem{lightFM}
M.~Kula, ``Metadata embeddings for user and item cold-start recommendations,''
  \emph{arXiv preprint arXiv:1507.08439}, 2015.

\end{thebibliography}
